\documentclass[10pt,twocolumn,letterpaper]{article}

\usepackage{cvpr}              

\definecolor{cvprblue}{rgb}{0.21,0.49,0.74}
\usepackage[pagebackref,breaklinks,colorlinks,allcolors=cvprblue]{hyperref}

\usepackage{xspace}
\usepackage{amssymb}
\usepackage{amsmath}
\usepackage{amsfonts}
\usepackage{wrapfig}
\usepackage{algorithm}
\usepackage{algorithmic}

\usepackage{booktabs}
\usepackage{multicol}
\usepackage{multirow}
\usepackage{makecell}
\usepackage{wrapfig}
\usepackage{tabularx}
\usepackage{adjustbox}

\usepackage{pifont}
\usepackage{color}
\usepackage{colortbl}
\usepackage[accsupp]{axessibility}

\definecolor{lightgray}{rgb}{0.8, 0.8, 0.8}
\definecolor{lgray}{rgb}{0.66, 0.66, 0.66}

\definecolor{lblu_tab}{RGB}{225, 235, 246}
\definecolor{orange_vitad}{RGB}{222, 131, 68}
\definecolor{blue_vitad}{RGB}{106, 153, 208}
\definecolor{trajectory_green}{RGB}{126, 171, 85}
\definecolor{trajectory_yellow}{RGB}{245, 194, 66}
\definecolor{tab_others}{RGB}{235, 235, 235}
\definecolor{tab_ours}{RGB}{225, 235, 246}

\definecolor{whit_tab}{RGB}{255, 255, 255}
\definecolor{gray_tab}{RGB}{246, 246, 246}
\definecolor{oran_tab}{RGB}{252, 242, 237}
\definecolor{blue_tab}{RGB}{227, 240, 251}

\newcommand{\et}[2]{${#1}^{\pm{#2}}$}
\newcommand{\etb}[2]{$\mathbf{{#1}}^{\pm{#2}}$}
\newcommand{\ets}[2]{$\underline{{#1}}^{\pm{#2}}$}

\newcommand{\cmark}{\ding{52}\xspace}%
\usepackage[accsupp]{axessibility}

\def\paperID{11113} 
\def\confName{CVPR}
\def\confYear{2025}

\title{TIMotion: Temporal and Interactive Framework for \\ Efficient Human-Human Motion Generation}

\author{
    Yabiao Wang$^{1,2}$\footnotemark[1],
    ~~~Shuo Wang$^{2}$\thanks{Equal contributions; author order is arbitrary.},
    ~~~Jiangning Zhang$^{2}$,
    ~~~Ke Fan$^{3}$, \\
    ~~~Jiafu Wu$^{2}$,
    ~~~Zhucun Xue$^{1}$,
    ~~~Yong Liu$^{1}$\thanks{Corresponding author.} 
    \\
    $^1$Zhejiang University~~~$^2$Youtu Lab, Tencent~~~$^3$Shanghai Jiao Tong University~~~\\
   \small {\textcolor{magenta}{https://aigc-explorer.github.io/TIMotion-page/}}\\
}

\begin{document}
\maketitle
\begin{abstract}
Human-human motion generation is essential for understanding humans as social beings. Current methods fall into two main categories: single-person-based methods and separate modeling-based methods. To delve into this field, we abstract the overall generation process into a general framework MetaMotion, which consists of two phases: temporal modeling and interaction mixing. For temporal modeling, the single-person-based methods concatenate two people into a single one directly, while the separate modeling-based methods skip the modeling of interaction sequences. The inadequate modeling described above resulted in sub-optimal performance and redundant model parameters. In this paper, we introduce TIMotion (Temporal and Interactive Modeling), an efficient and effective framework for human-human motion generation. Specifically, we first propose Causal Interactive Injection to model two separate sequences as a causal sequence leveraging the temporal and causal properties. Then we present Role-Evolving Scanning to adjust to the change in the active and passive roles throughout the interaction. Finally, to generate smoother and more rational motion, we design Localized Pattern Amplification to capture short-term motion patterns.
Extensive experiments on InterHuman and Inter-X demonstrate that our method achieves superior performance.


\end{abstract}    
\section{Introduction} \label{sec:introduction}
In the field of generative computer vision, human motion generation has significant implications for computer animation~\cite{parent2012computer, magnenat1985computer}, game development~\cite{urbain2010introduction, bethke2003game}, and robotic control~\cite{saridis1983intelligent, wang2023towards, wang2024sqdmap}. In recent years, there have been remarkable advancements in human motion generation, driven by various user-specified conditions such as action categories~\cite{petrovich2021action, guo2020action2motion}, speeches~\cite{ao2022speech, habibie2022speech}, and natural language prompts~\cite{pinyoanuntapong2024mmm, guo2024momask}.
Among these, many approaches leveraging large language models~\cite{achiam2023gpt} and diffusion models~\cite{zhao2024wavelet} have yielded impressive results in generating realistic and diverse motions, benefiting from their powerful modeling capabilities.
Despite this progress, most existing methods are designed primarily for single-person scenarios, thereby neglecting a crucial element of human motion: the complex and dynamic interactions between individuals.
We address the challenge of generating two-person motion by fully utilizing the temporal and interactive dynamics between the individuals.

\begin{figure*}[ht]
  \centering
   \includegraphics[width=1.0\linewidth]{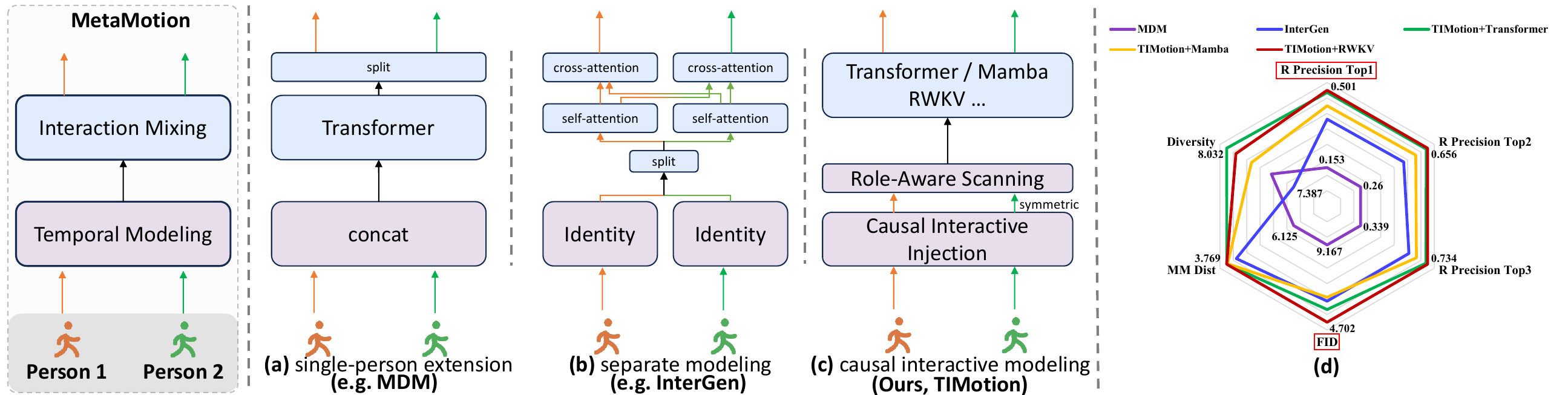}
   \caption{
   \textbf{MetaMotion and performance of MetaMotion-based models on InterHuman validation set.} We abstract the MetaMotion concept that illustrates the intrinsic properties of human-human motion generation in the interaction process. (a) and (b) show the two types of methods currently, and (c) shows our method TIMotion, LPA refers to the Localized Pattern Amplification. In (d) we compare the performance of the different methods on the InterHuman dataset.
}
   \label{fig:teaser}
\end{figure*}

To better explore human-human motion generation, we abstract a general framework \emph{MetaMotion}, as shown on the left of ~\cref{fig:teaser}, which consists of two phases: temporal modeling and interaction mixing. Previous approaches have prioritized interaction mixing over temporal modeling and can be divided into two primary categories: single-person-based methods and separate modeling-based methods. As shown in~\cref{fig:teaser}(a), the single-person-based methods (\eg MDM~\cite{GuyTevet_MDM}) concatenate two individuals into one, which is then fed into the existing single-person motion generation module, \eg DiT. And the separate modeling-based methods (\eg InterGen~\cite{liang2024intergen}), as shown in~\cref{fig:teaser}(b), model two individuals individually and then extract motion information from both themselves and each other, using self-attention and cross-attention mechanisms, respectively.
Following the general logic of \emph{MetaMotion}, we introduce the Temporal and Interactive Framework, as depicted in~\cref{fig:teaser}(c), which models the human-human causal interactions.
In fact, this effective temporal modeling method can simplify the design of the interaction mixing module and reduce the number of learnable parameters.

In this paper, we design an effective temporal modeling method and introduce TIMotion, a temporal and interactive framework for efficient human-human motion generation.
Specifically, we first propose Causal Interactive Injection to model the two single-person motion sequences as a causal interaction sequence, according to the temporal causal properties of motion sequences. Second, since active and passive roles are not static in interactions, we propose Role-Evolving Mixing to make two humans act in both active and passive roles. Then the network can dynamically adjust the roles of the two humans based on the text's semantics and the motion's context.
Finally, we propose Localized Pattern Amplification, which captures short-term motion patterns for each individual separately, resulting in smoother and more logical motion generation.
Our proposed framework can be well adapted to different interaction-mixing modules, including Transformer, RWKV, and  Mamba.

Our contributions can be summarized as follows:

\begin{itemize}
\item We conceptualize human-human motion generation within a general framework MetaMotion and design an innovative method. Our proposed framework, TIMotion, is versatile enough to integrate with various interaction-mixing modules (\eg Transformer, RWKV, Mamba) and reduces the number of parameters of these modules.
\item To utilize the temporal and causal properties, we propose Causal Interactive Injection to model two separate motion sequences as a unified causal sequence. In addition, we introduce Role-Evolving Scanning to accommodate shifts between the active and passive roles during interactions. Moreover, we also design Localized Pattern Amplification to capture short-term motion patterns effectively.
\item We perform extensive experiments on the benchmark human-human motion generation datasets: InterHuman and Inter-X. The results demonstrate the effectiveness and generalizability of our proposed methods. 
\end{itemize}
\section{Related Work} \label{sec:related_work}
\paragraph{Single-Person Human Motion Generation.}
Creating human motion is vital for applications such as 3D modeling and robot manipulation. The primary approach, known as the Text-to-Motion task, involves learning a unified latent space for both language and motion. 

Autoencoders have been widely adopted in human motion generation. MotionCLIP~\cite{tevet2022motionclip} effectively integrates semantic knowledge from CLIP~\cite{radford202clip} into the human motion manifold. TEMOS~\cite{petrovich2022temos} and T2M~\cite{guo2022t2m} combine a Transformer-based VAE with a text encoder to generate motion sequences based on text descriptions. AttT2M~\cite{zhong2023attt2m} and TM2D~\cite{gong2023tm2d} integrate a spatial-temporal body-part encoder into VQ-VAE~\cite{van2017vqvae} to improve the learning of discrete latent space. T2M-GPT~\cite{zhang2023t2mgpt} redefines text-driven motion generation as a next-index prediction task and proposes a framework based on  VQ-VAE and Generative Pretrained Transformer (GPT) for motion generation.

Recently, diffusion-based generative modeling has been gaining attention. MotionDiffuse~\cite{zhang2024motiondiffuse} exhibits several desirable properties, such as probabilistic mapping, realistic synthesis, and multi-level manipulation. MDM~\cite{tevet2209mdm} aims to predict motion directly and incorporates a geometric loss to boost the model's performance. Instead of employing a diffusion model to link raw motion sequences with conditional inputs, MLD~\cite{chen2023mld} further utilizes the latent diffusion model to reduce the training and inference costs substantially. ReMoDiffuse~\cite{zhang2023remodiffuse} introduces an improvement mechanism based on dataset retrieval to enhance the denoising process of Diffusion. 

\begin{figure*}[h]
  \centering
   \includegraphics[width=1.0\linewidth]{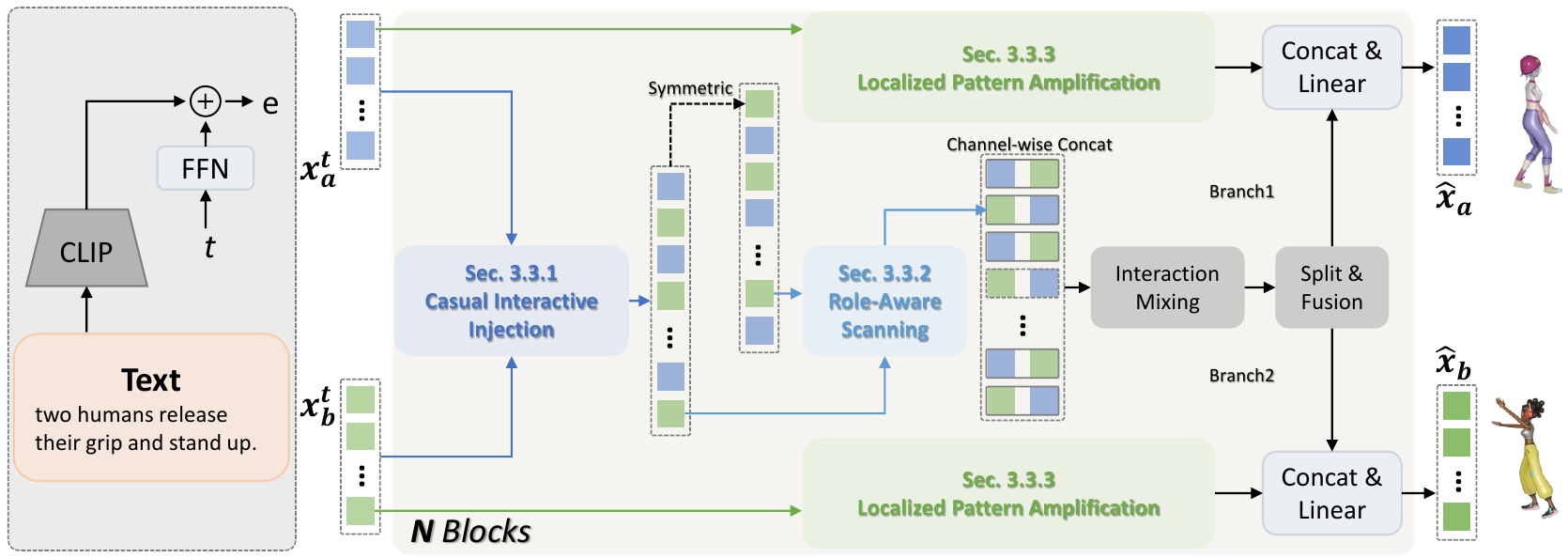}
   \caption{\textbf{The overall framework of our TIMotion.} We contribute three primary technical designs. First, we propose Causal Interactive Injection to utilize the temporal properties of motion sequences. Then we present Role-Evolving Mixing to adjust to the ever-evolving roles during interaction. Finally, we design Localized Pattern Amplification to capture short-term motion patterns.
}
   \label{fig:network}
\end{figure*}

\paragraph{Multi-Person Human Motion Generation.}
As multi-person motion synthesis involves the interactive dynamics between multiple individuals, it is more challenging than single-person motion generation. Early work typically relied on motion graphs and momentum-based inverse kinematics to model human joints. Recently, ComMDM~\cite{shafir2023commdm} finetunes a pre-trained text-to-motion diffusion model using a small scale of two human motions. RIG~\cite{tanaka2023rig} converts the text of asymmetric interactions into both active and passive voice to maintain consistent textual context for each individual. To complete the interaction process by manipulating the joints of two individuals to specific positions, InterControl~\cite{wang2023intercontrol} employs a Large Language Model to generate movement plans for the two individuals by designing prompts. InterGen~\cite{liang2024intergen} introduces a large-scale, text-annotated two-person motion dataset. Building on this dataset, it proposes a diffusion model with shared weights and multiple regularization losses. FreeMotion~\cite{fan2024freemotion} proposes to decouple the process of conditional motion generation and support the number-free motion synthesis.

\section{Method} \label{sec:method}

\subsection{Motion Diffusion Model}

Current methods usually use diffusion models~\cite{ho2020denoising} for motion generation. For the Interhuman dataset, a non-canonical representation~\cite{liang2024intergen} is typically used. The representation is formulated as: $x^i=[\textbf{j}^{p}_{g} , \textbf{j}^{v}_{g} , \textbf{j}^r, \textbf{c}^f ]$, where the $i$-th motion state $x^i$ is defined as a collection of global joint positions $\textbf{j}^{p}_{g}\in \mathbb{R}^{3N_j}$, velocities $\textbf{j}^{v}_{g}\in \mathbb{R}^{3N_j}$ in the world frame, 6D representation of local rotations $\textbf{j}^{r}\in \mathbb{R}^{6N_j}$ in the root frame, and binary foot-ground contact $\textbf{c}^f \in \mathbb{R}^{4}$. 
Our goal is to train a model parameterized by $\theta$ to approximate the human interactive motion data
 distribution $p(\mathbf{x_0})$. 

\noindent \textbf{Diffusion Process.} Following a specified schedule $\beta_t \in (0, 1)$, models incrementally degrade input data  $\mathbf{x}_0 \sim p(\mathbf{x}_0)$, ultimately transforming the data distribution into an isotropic Gaussian over $T$ steps. Each diffusion transition can be considered as
\begin{align}
q\left(\mathbf{x}_{t} \mid \mathbf{x}_{t-1}\right) & = \mathcal{N}\left(\mathbf{x}_{t} ; \sqrt{1-\beta_{t}} \mathbf{x}_{t-1}, \beta_{t} \mathbf{I}\right),
\end{align}
where the full diffusion process can be written as
\begin{align}
q\left(\mathbf{x}_{1: T} \mid \mathbf{x}_{0}\right) & = \prod_{1 \leq t \leq T} q\left(\mathbf{x}_{t} \mid \mathbf{x}_{t-1}\right).
\end{align}
\noindent \textbf{Denoising Process.} 
In the denoising process, models are trained to reverse the diffusion procedure, enabling them to transform random noise into real data distribution during inference.
The denoising process can be written as
\begin{eqnarray}
& p_{\theta}\left(\mathbf{x}_{t-1} \mid \mathbf{x}_{t}\right) = \mathcal{N}\left(\mathbf{x}_{t-1} ; \mu_{\theta}\left(\mathbf{x}_{t}, t\right), \Sigma_{\theta}\left(\mathbf{x}_{t}, t\right)\right) \nonumber\\ &= \mathcal{N}\left(\mathbf{x}_{t-1} ; \frac{1}{\sqrt{\alpha_{t}}}\left(\mathbf{x}_{t}-\frac{\beta_{t}}{\sqrt{1-\bar{\alpha}_{t}}} \mathbf{\epsilon} \right),
\frac{1-\bar{\alpha}_{t-1}}{1-\bar{\alpha}_{t}} \beta_{t}\right),
\end{eqnarray}
where $\mathbf{\epsilon} \sim \mathcal{N}(\mathbf{0},\mathbf{I})$, $\alpha_{t}=1-\beta_{t}$, $\bar{\alpha}_{t}=\prod_{i=1}^{t} \alpha_{i}$ and $\theta$ denotes parameters of the network learning to denoise. The training objective is to maximize the likelihood of observed data $p_{\theta}\left(\mathbf{x}_{0}\right)=\int p_{\theta}\left(\mathbf{x}_{0: T}\right) d \mathbf{x}_{1: T}$, by maximizing its evidence lower bound (ELBO), which effectively aligns the true denoising model $q\left(\mathbf{x}_{t-1} \mid \mathbf{x}_{t}\right)$ with the parameterized $p_{\theta}\left(\mathbf{x}_{t-1} \mid \mathbf{x}_{t}\right)$. During training, the goal of the denoising network $\mathbf{\epsilon}_\theta(.)$ is to reconstruct $\mathbf{x}_0$ given any noised input $\mathbf{x}_t$, by predicting the added noise $\epsilon \sim \mathcal{N}(\mathbf{0},\mathbf{I})$ via minimizing the noise prediction error 
\begin{align}
\mathcal{L}_{t} & = \mathbb{E}_{\mathbf{x}_{0}, \epsilon }\left[\left\|\mathbf{\epsilon}-\mathbf{\epsilon}_{\theta}\left(\sqrt{\bar{\alpha}_{t}}\mathbf{x}_0+\sqrt{1-\bar{\alpha}_{t}} \mathbf{\epsilon}, t\right)\right\|^{2}\right]. \label{equ:noise_loss}
\end{align}
To condition the model on additional context information $\mathbf{c}$, \textit{e.g.}, text, we inject $\mathbf{c}$ into $\mathbf{\epsilon}_\theta(.)$ by replacing 
$\mu_{\theta}\left(\mathbf{x}_{t}, t\right)$ and $\Sigma_{\theta}\left(\mathbf{x}_{t}, t\right)$ with $\mu_{\theta}\left(\mathbf{x}_{t}, t, \mathbf{c}\right)$ and $\Sigma_{\theta}\left(\mathbf{x}_{t}, t,\mathbf{c}\right)$.

\subsection{MetaMotion}
We present the core concept ``MetaMotion" for human-human motion generation at first. As shown in~\cref{fig:teaser}, we abstract the process of human-human motion generation into two phases: temporal modeling and interaction mixing.

Specifically, two single-person sequences go through the temporal modeling module to get the input sequence $X$. Then, the input sequence is fed to the interaction-mixing module and this process can be expressed as
\begin{equation}
  Y = {\rm InteractionMixing} (X),
  \label{eq:0}
\end{equation}
where ${\rm InteractionMixing}$ is usually the transformer structure including self-attention and cross-attention. Notably, ${\rm InteractionMixing}$ can also be some emerging structures, \eg, Mamba~\cite{gu2023mamba}, RWKV~\cite{peng2023rwkv}.


\subsection{TIMotion}

For human-human motion generation, most current methods are based on extensions of single-person motion generation. Some design-specific approach (\eg InterGen~\cite{liang2024intergen}) considers the interaction between two persons in the interaction mixing module. 
However, current methods ignore the importance of temporal modeling, resulting in suboptimal performance in generating long-sequence motions and multi-person interactions.

In this work, we propose TIMotion, as illustrated in ~\cref{fig:network}, an effective temporal modeling approach that can be applied to different interaction-mixing structures, enabling better handling the human-human interactions.
TIMotion consists of three key technical designs: (1) Causal Interactive Injection, which identifies the causal relationships between human-human interactions; (2) Role-Evolving Scanning, which adapts to shifts between active and passive roles during interactions; (3) Localized Pattern Amplification, designed to handle short-term sequences better.


\subsubsection{Causal Interactive Injection} \label{sec:cii}

Perception of ego-motion and interaction between two persons are two important elements of human-human motion generation. Considering the causal properties of motion, we propose Causal Interactive Injection, a temporal modeling approach to simultaneously achieve both perceptions of ego-motion and interactions with each other.


Specifically, we denote two single-person motion sequences as \{$x_a$, $x_b$\}, where $x_a=\{x_a^j\}^L_{j=1}$ and $x_b=\{x_b^j\}^L_{j=1}$ are respective sequences of motion, and $L$ is the length of the sequence. 
Since the motions of two individuals at the current time step are jointly determined by their motions at previous time steps, we model the two single-person motion sequences as a causal interaction sequence $x_{cii} = \{x^{j // 2}_k\}_{j=1}^{2L}$, the symbol \texttt{\small $//$} denotes division followed by rounding up and $k$ can be acquired as follows:
\begin{equation}
  k = \begin{cases}
    a,& j \; \% \; 2 = 1.  \\
    b,& j \; \% \; 2 = 0. 
  \end{cases} 
  \label{eq:1}
\end{equation}

Then we can inject them into the interaction-mixing module and separate the motion embeddings of the two individuals from the outputs according to ~\cref{eq:1}.

\begin{figure}[h]
  \centering
   \includegraphics[width=0.9\linewidth]{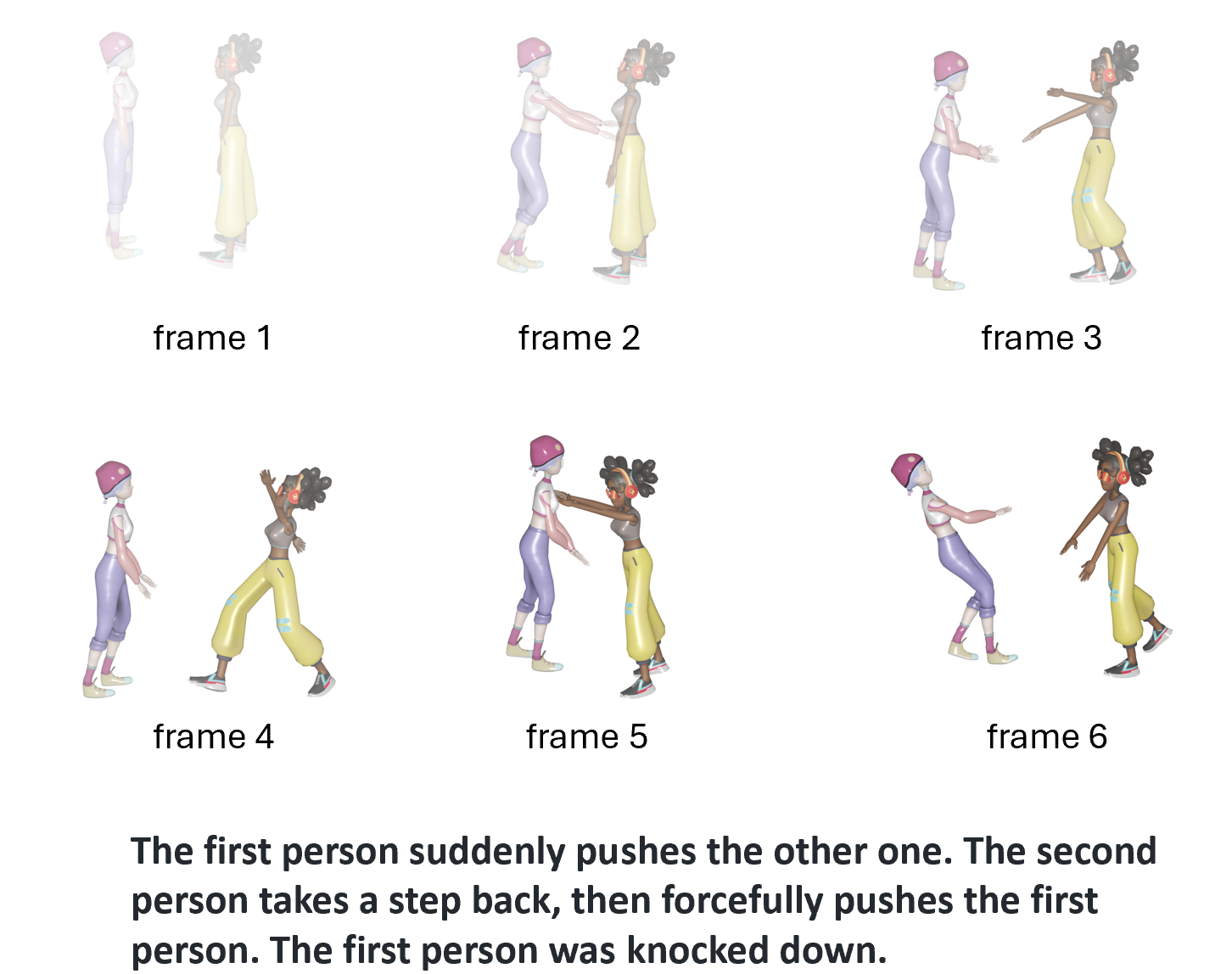}
   \caption{Illustration of changing active and passive roles. The first person acts as the active role in the early stages, and as time progresses, the other person becomes the active role of the motion.
}
   \label{fig:method2}
\end{figure}

\subsubsection{Role-Evolving Scanning}
As noted in~\cite{cai2024digital}, human interactions inherently follow a certain order, \eg, ``shaking hands" typically begins with one person extending a hand first, which means that interactive motion can be categorized as active or passive. Some methods~\cite{tanaka2023rig} split the text description into active and passive voices. However, the ``active" and ``passive" roles are not static in a text description. As shown in~\cref{fig:method2}, these roles constantly swap between characters as the interaction progresses. To minimize redundant text preprocessing and adapt to the ongoing shifts in roles, we design Role-Evolving Scanning, which is both efficient and effective.

For the causal interaction sequence $x$ defined in Causal Interactive Injection (\cref{sec:cii}), it is obvious that $x_a$ and $x_b$ represent the active sequence and passive sequence, respectively. However, the above assumption of active and passive motions is not always in line with the real order. To cope with the changing roles, we re-model the interactive motion sequences as a symmetric causal interaction sequence $x_{sym\_cii} = \{x^{j // 2}_{k^{\prime}}\}_{j=1}^{2L}$, where the symbol $//$ is defined as division followed by rounding up and $k^{\prime}$ is obtained by exchanging $a$ and $b$ in~\cref{eq:1}.

Given that the causal interaction sequence $x_{cii} \in \mathbb{R}^{2L \times C}$ and the symmetric causal interaction sequence $x_{sym\_cii} \in \mathbb{R}^{2L \times C}$, where $L$ is the length of the sequence and $C$ is the dimension of motion embedding, we can acquire the final interaction sequence $X \in \mathbb{R}^{2L \times 2C}$ through Role-Evolving Scanning as:
\begin{equation}
  X = {\rm Concat}(x_{cii}, \; x_{sym\_cii}).
  \label{eq:analysis}
\end{equation}

Then the sequence $X$ is fed to the interaction-mixing module. 
After obtaining the output $Y \in \mathbb{R}^{2L \times 2C}$ through~\cref{eq:0}, we first divide it at the channel level into causal interaction embeddings and symmetric causal interaction embeddings. According to $x_{cii}$ and$x_{sym\_cii}$, we can obtain the two individuals' splitting embeddings twice. Then we merge the embeddings of the two individuals to obtain the final global motion embeddings $y_a^{g} \in \mathbb{R}^{L \times C}$ and $y_b^{g} \in \mathbb{R}^{L \times C}$. The overall process is as follows:
\begin{equation}
\begin{aligned}
  y_{a1}, \; y_{b1} &= {\rm Split}(Y[:, \; :C]),  \\
  y_{b2}, \; y_{a2} &= {\rm Split}(Y[:, \; C:]),   \\
 y_a^{g} &= y_{a1} \oplus y_{a2}, \\
 y_b^{g} &= y_{b1} \oplus y_{b2}, 
\end{aligned}
\label{eq:6}
\end{equation}

where $\oplus$ denotes element-wise sum.

By utilizing Role-Evolving Mixing to make two humans act as both active and passive roles, the network can dynamically adjust the roles of the two humans based on the semantics of the text and the context of the motion.

\subsubsection{Localized Pattern Amplification}
Transformers and RNNs excel at global modeling and capturing long-range dependencies but tend to overlook local semantic information~\cite{he2024mambaad}. Additionally, Causal Interactive Injection and Role-Evolving Mixing mainly model the overall motion based on the causality of the interactions, but neglect to focus on the localized motion patterns of the single person. To address this issue, we propose Localized Pattern Amplification, which summarizes short-term motion patterns for each person individually while generating smoother and more rational motion.

Specifically, we utilize 1-D convolution layers and the residual structure to realize Localized Pattern Amplification. Given that the condition embedding $e$ and two single-person motion sequences $x_a$ and $x_b$, the local motion embedding for $x_a$ can be expressed as: 
\begin{align}
    x_a^l &= {\rm Conv_3}({\rm AdaLN}(x_a, \; e)), \notag \\
    y_a &= {\rm Conv_1}({\rm AdaLN}(x_a^l, \; e)), \\
    y_a^l &= x_a + y_a, \notag
\end{align}
where $\rm Conv_k$ denotes the 1-D convolution with kernel size k and ${\rm AdaLN}$ is the adaptive layer normalization used in~\cite{zhang2024motiondiffuse}. The process for $x_b$ is the same as for $x_a$ and they share network weights.

Once we obtain the outputs \{$y_a^{g}$, $y_b^{g}$\} through~\cref{eq:6} and the output \{$y_a^l$, $y_b^l$\} from the convolution block, global embeddings and local embeddings are aggregated through concatenation along the channel dimension, followed by a linear layer to restore the original channels, resulting in the final outputs \{$y_a^{final}$, $y_b^{final}$\}: 
\begin{align}
    y_a^{final} &= {\rm Linear}({\rm Concat}(y_a^g, \; y_a^l)), \notag \\
    y_b^{final} &= {\rm Linear}({\rm Concat}(y_b^g, \; y_b^l)).
\end{align}
Then the final outputs are fed into the next encoder block or the final decoder layer. 


\subsubsection{Objective Function}
We use the same loss functions as InterGen~\cite{liang2024intergen}, including the diffusion loss $\mathcal{L}_{simple}$, the foot contact loss $\mathcal{L}_{foot}$, joint velocity loss $\mathcal{L}_{vel}$, the bone length loss $\mathcal{L}_{BL}$, the masked joint distance map loss $\mathcal{L}_{DM}$, and the relative orientation loss $\mathcal{L}_{RO}$. For more details about the losses, we refer readers to InterGen. 
Finally, the overall loss is defined as: 
\begin{align}
  \mathcal{L}_{motion} = \mathcal{L}_{simple} + \lambda_{vel }\mathcal{L}_{vel} + \lambda_{foot }\mathcal{L}_{foot} \notag \\ + \lambda_{BL }\mathcal{L}_{BL} 
  + \lambda_{DM }\mathcal{L}_{DM} + \lambda_{RO }\mathcal{L}_{RO},
\end{align}
where the hyper-parameters $\lambda_{vel},\lambda_{foot},\lambda_{BL},\lambda_{DM},\lambda_{RO}$ are the same as InterGen.


\section{Experiments} \label{sec:exp}
\begin{table*}[t]
    \centering
    \scalebox{0.88}{
    \begin{tabular}{l c c c c c c c}
    \toprule
    \multirow{2}{*}{Methods}  & \multicolumn{3}{c}{R Precision$\uparrow$} & \multirow{2}{*}{FID$\downarrow$} & \multirow{2}{*}{MM Dist$\downarrow$} & \multirow{2}{*}{Diversity$\rightarrow$} &\multirow{2}{*}{MModality$\uparrow$}\\

    \cline{2-4}
     ~ & Top 1 & Top 2 & Top 3 \\

     \cline{1-8}
     Real & \et{0.452}{.008} & \et{0.610}{.009} & \et{0.701}{.008} & \et{0.273}{.007} & \et{3.755}{.008} & \et{7.948}{.064} & - \\
    \midrule
    TEMOS~\cite{petrovich2022temos} & \et{0.224}{.010} & \et{0.316}{.013} & \et{0.450}{.018} & \et{17.375}{.043} & \et{5.342}{.015} & \et{6.939}{.071} & \et{0.535}{.014} \\ 
        T2M~\cite{guo2022t2m} & \et{0.238}{.012} & \et{0.325}{.010} & \et{0.464}{.014} & \et{13.769}{.072} & \et{4.731}{.013} & \et{7.046}{.022} & \et{1.387}{.076} \\   
        MDM~\cite{tevet2209mdm} & \et{0.153}{.012} & \et{0.260}{.009} & \et{0.339}{.012} & \et{9.167}{.056} & \et{6.125}{.018} & \et{7.602}{.045} & \etb{2.355}{.080}  \\
        ${\rm ComMDM^*}$~\cite{shafir2023commdm} & \et{0.067}{.013} & \et{0.125}{.018} & \et{0.184}{.015} & \et{38.643}{.098} & \et{13.211}{.013} & \et{3.520}{.058} & \et{0.217}{.018}  \\
        ComMDM~\cite{shafir2023commdm} & \et{0.223}{.009} & \et{0.334}{.008} & \et{0.466}{.010} & \et{7.069}{.054} & \et{5.212}{.021}  & \et{7.244}{.038} & \et{1.822}{.052} \\

       RIG~\cite{tanaka2023rig} & \et{0.285}{.010} & \et{0.409}{.014} & \et{0.521}{.013} & \et{6.775}{.069} & \et{4.876}{.018}  & \et{7.311}{.043} & \et{2.096}{.065}  \\

        InterGen~\cite{liang2024intergen} & \et{0.371}{.010} & \et{0.515}{.012} & \et{0.624}{.010} & \et{5.918}{.079} & \et{5.108}{.014} & \et{7.387}{.029} & \ets{2.141}{.063}  \\

        
    \cline{1-8}
        \textbf{TIMotion+Transformer(ours)} & \ets{0.491}{.005} & \ets{0.648}{.004} & \ets{0.724}{.004} & \ets{5.433}{.080} & \ets{3.775}{.001} & \ets{8.032}{.030} & \et{0.952}{.032}  \\
        \textbf{TIMotion+Mamba(ours)} & \et{0.431}{.004} & \et{0.586}{.005} & \et{0.668}{.004} & \et{6.142}{.059} & \et{3.800}{.001}  & \et{7.793}{.024} & \et{0.837}{.036}  \\
        \textbf{TIMotion+RWKV(ours)} & \etb{0.501}{.005} & \etb{0.656}{.006} & \etb{0.734}{.006} & \etb{4.702}{.069} & \etb{3.769}{.001}& \etb{7.943}{.034} & \et{1.005}{.020}  \\
        
    \bottomrule
    \end{tabular}
    }
    \caption{\textbf{Quantitative evaluation on the InterHuman~\cite{liang2024intergen} test set.} We run all the evaluations 20 times. $\pm$ indicates a 95\% confidence interval. \textbf{Bold} indicates the best result, while \underline{underline} refers to the second best. ComMDM$^{*}$ indicates the ComMDM model fine-tuned in the original few-shot setting with 10 training samples and ComMDM indicates fine-tuned on the entire InterHuman training set.}
    \label{tab:main_interhuman}

\end{table*}

\begin{table*}[t]
    \centering
    \scalebox{0.73}{
    \begin{tabular}{l l c c c c c c c}
    \toprule
    \multirow{2}{*}{Interaction Mixing}  & \multirow{2}{*}{Temporal Modeling}  &\multicolumn{3}{c}{R Precision$\uparrow$} & \multirow{2}{*}{FID$\downarrow$} & \multirow{2}{*}{MM Dist$\downarrow$} & \multirow{2}{*}{Diversity$\rightarrow$} &\multirow{2}{*}{MModality$\uparrow$}\\

    \cline{3-5}
     ~ & ~& Top 1 & Top 2 & Top 3 \\

    \midrule
    \multirow{3}{*}{Transformer~\cite{vaswani2017transformer}} &single-person extension& \et{0.395}{.004} & \et{0.535}{.004} & \et{0.615}{.004} & \et{8.028}{.099} & \et{3.820}{.001} & \et{7.687}{.022} & \et{0.852}{.024} \\ 
        ~ & separate modeling &\et{0.371}{.010} & \et{0.515}{.012} & \et{0.624}{.010} & \et{5.918}{.079} & \et{5.108}{.014} & \et{7.387}{.029} & \etb{2.141}{.063} \\   
        ~ & \textbf{TIMotion (Ours)} & \etb{0.491}{.005} & \etb{0.648}{.004} & \etb{0.724}{.004} & \etb{5.433}{.080} & \etb{3.775}{.001} & \etb{8.032}{.030} & \et{0.952}{.032}  \\

        \cline{1-9}
        
        \multirow{3}{*}{Mamba~\cite{gu2023mamba}} &single-person extension & \et{0.368}{.005} & \et{0.513}{.005} & \et{0.597}{.006} & \et{7.232}{.081} & \et{3.825}{.001} & \et{7.864}{.024} & \etb{0.870}{.020}  \\
        ~ & separate modeling & \et{0.420}{.005} & \et{0.569}{.004} & \et{0.650}{.003} & \et{7.221}{.097} & \et{3.803}{.001}  & \etb{7.932}{.035} & \et{0.855}{.020} \\
       ~& \textbf{TIMotion (Ours)} & \etb{0.431}{.004} & \etb{0.586}{.005} & \etb{0.668}{.004} & \etb{6.142}{.059} & \etb{3.800}{.001}  & \et{7.793}{.024} & \et{0.837}{.036} \\

        \cline{1-9}

        \multirow{3}{*}{RWKV~\cite{peng2023rwkv}} &single-person extension & \et{0.425}{.004} & \et{0.576}{.004} & \et{0.656}{.003} & \et{9.181}{.096} & \et{3.801}{.001} & \et{7.679}{.022} & \et{0.846}{.026}  \\
        ~ & separate modeling & \et{0.465}{.005} & \et{0.603}{.005} & \et{0.689}{.004} & \et{5.943}{.079} & \et{3.790}{.001}  & \et{7.787}{.032} & \et{0.859}{.036} \\
       ~& \textbf{TIMotion (Ours)} & \etb{0.501}{.005} & \etb{0.656}{.006} & \etb{0.734}{.006} & \etb{4.702}{.069} & \etb{3.769}{.001}& \etb{7.943}{.034} & \etb{1.005}{.020}  \\
        
    \bottomrule
    \end{tabular}
    }
    \caption{\textbf{Comparison of different temporal modeling approaches on different interaction mixing structures.} Our proposed causal interactive modeling is able to adapt to different interaction mixing architectures and outperforms the other two ways.}
    \label{tab:new_ablation}

\end{table*}

\begin{figure*}[h]
  \centering
   \includegraphics[width=1.0\linewidth]{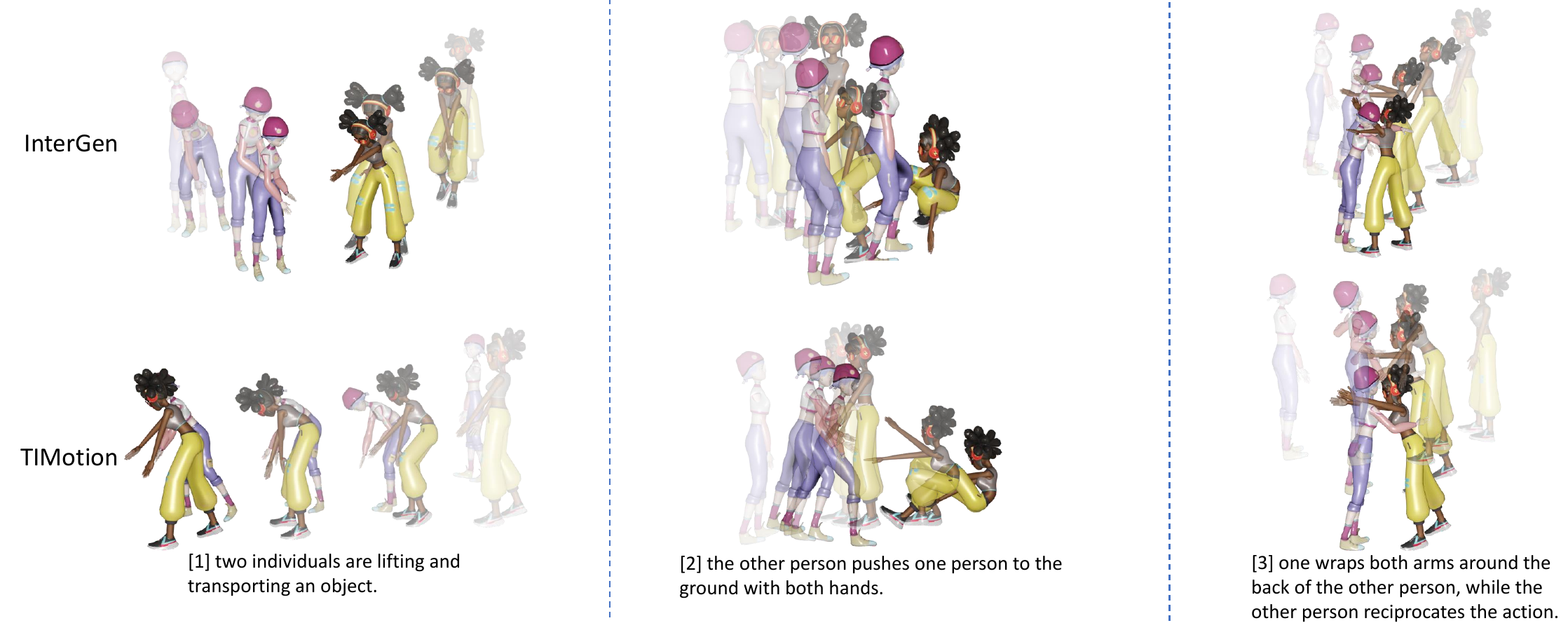}
   \caption{Qualitative comparison with Intergen on human-human motion generation. Darker color indicates later frames.
   The sequences generated by TIMotion are more consistent with the text description.
}
   \label{fig:vis}
\end{figure*}

\subsection{Experimental Setup}
\noindent\textbf{Datasets.} We assess our proposed framework using the InterHuman~\cite{liang2024intergen} and Inter-X~\cite{xu2024interx} dataset. InterHuman is the first dataset to incorporate text annotations for two-person motions. This dataset includes 6,022 motions spanning various categories of human actions and is labeled with 16,756 unique descriptions made up of 5,656 distinct words. Inter-X is the largest human-human interaction dataset with diverse interaction patterns. It includes about 11k interaction sequences and more than 8.1M frames.

\noindent\textbf{Metrics.} 
We employ the same evaluation metrics as InterGen~\cite{liang2024intergen}, which are as follows: (1) \emph{Frechet Inception Distance (FID)}. (2) \emph{R-Precision}. (3) \emph{Diversity}. (4) \emph{Multimodality (MModality)}. (5) \emph{Multi-modal distance (MM Dist)}. Detailed explanations and calculations of the metrics can be found in the \textbf{supplementary materials}.

\subsubsection{Implementation Details.}
We use a frozen CLIP-ViT-L/14 model as the text encoder. The dimension of motion embedding is set to 512. During training, the number of diffusion timesteps is set to 1000, and we employ the DDIM~\cite{song2020diffusion} sampling strategy with 50 timesteps and $\eta = 0$. The cosine noise level schedule~\cite{nichol2021cosnoise} and classifier-free guidance~\cite{ho2022classifier} are adopted, with 10\% of random CLIP embeddings set to zero during training and a guidance coefficient of 3.5 during sampling.
All the models are trained using the AdamW~\cite{loshchilov2017adam} optimizer with betas of (0.9, 0.999), a weight decay of $2\times10^{-5}$, a maximum learning rate of $10^{-4}$, and a cosine learning rate schedule with 10 linear warm-up epochs. To balance the contributions of different loss terms, we set $\lambda_{vel} = 30$, $\lambda_{foot} = 30$, $\lambda_{BL} = 10$, $\lambda_{DM} = 3$, $\lambda_{RO} = 0.01$ and $\lambda_{reg} = 1$ in all experiments, as same as InterGen.
We train our diffusion denoisers with a batch size of 256 for 1500 epochs on 8 Nvidia L40S GPUs.

\subsection{Comparisons with State-of-the-arts}
\subsubsection{Quantitative Results.} 
Following established practices~\cite{liang2024intergen}, each experiment is conducted 20 times, and the reported metric values represent the mean with a 95\% statistical confidence interval. The results on InterHuman are shown in~\cref{tab:main_interhuman}. In comparison to existing methods, our TIMotion achieves competitive results on three different interaction mixing architectures: transformer, mamba, and RWKV. Equipped with RWKV, TIMotion achieves 4.702 FID and 0.501 Top1 R precision, setting new state-of-the-art (SoTA) for the competitive InterHuman benchmark.
The results on Inter-X~\cite{xu2024interx} can be found in the \textbf{supplementary materials}.

\subsubsection{Qualitative Comparisons.}
In ~\cref{fig:vis}, we qualitatively compare the InterGen and our TIMotion. It can be seen that sequences generated by TIMotion are more consistent with the description.

\subsection{Effectiveness of TIMotion}
To validate the effectiveness of our proposed causal interactive modeling (TIMotion), we compare different temporal modeling approaches on three different structures, respectively. As shown in~\cref{tab:new_ablation}, single-person extension performs poorly, indicating that it does not model two-person interactions well. In contrast, separate modeling leverages the ability of the interaction mixing module to model the two-person interactions but achieves sub-optimal performance on the Transformer and mamba architectures. Our proposed approach is able to adapt to different interaction mixing architectures and outperforms the other two approaches.

\subsection{Ablation Study and Analysis}
In this section we conduct comprehensive ablation studies to investigate the effectiveness of the key components in TIMotion, thereby providing a deeper insight into our approach. All experiments are performed on the InterHuman dataset. Unless otherwise stated, we use RWKV as the interaction mixing structure for the following ablations. More theoretical analyses are in the \textbf{supplementary materials}.

\begin{table}[!h]
    \centering
    \scalebox{0.88}{
    \begin{tabular}{ccc|ccc}
        \toprule
         \multirow{2}{*}{CII} & \multirow{2}{*}{RES}& \multirow{2}{*}{LPA}  & R Precision&  \multirow{2}{*}{FID $\downarrow$} & Params\\

           ~& ~ &  ~&Top 1$\uparrow$ &~&(M)   \\
           
         \midrule
           & &  &\et{0.371}{.010} &\et{5.918}{.079} &182 \\
          \cmark & & & \et{0.478}{.005} &\et{5.410}{.069} &144 \\
          \cmark & \cmark & & \et{0.494}{.006} &\et{5.019}{.077} &115   \\
          \cmark & \cmark & \cmark & \etb{0.501}{.005} &\etb{4.702}{.069}  &127 \\
        \bottomrule
    \end{tabular}
    }
    \caption{\textbf{Ablation studies on the effectiveness of each component in TIMotion.} ``CII'' denotes Causal Interactive Injection, ``RES'' denotes Role-Evolving Scanning, and ``LPA'' denotes Localized Pattern Amplification.}
    \label{tab:main_ablations}
\end{table}

\subsubsection{Main Ablations}
    To understand the contribution of each component to the final performance, we incrementally add the proposed modules based on the baseline InterGen~\cite{liang2024intergen} and present the results in \cref{tab:main_ablations}. Initially, we replace self-attention and cross-attention in the transformer with Causal Interactive Injection (CII), which effectively improves both R-Precision and FID while reducing the number of parameters. Next, we apply Role-Evolving Scanning (RES) and yield gains in both R-Precision and FID. It is worth noting that to keep the input feature dimension of RWKV constant, we reduce the dimension of the motion embedding to half of the original, thus also effectively reducing the number of parameters. Finally, when all three methods are applied together, the R-Precision achieves 0.501 and FID achieves 4.702.


\begin{figure*}[h]
  \centering
   \includegraphics[width=0.9\linewidth]{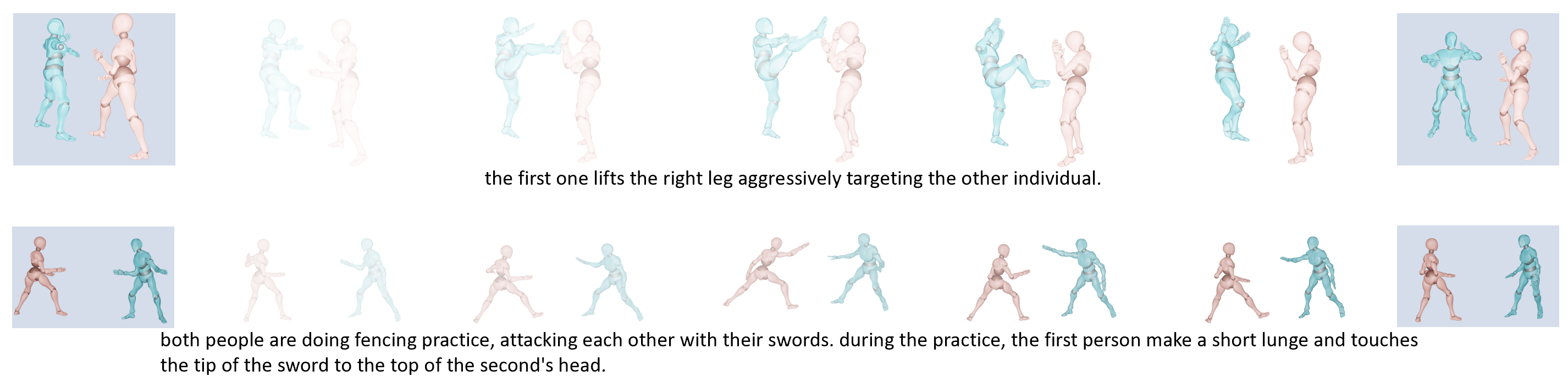}
   \caption{\textbf{Qualitative results on the motion in-betweening task.}  The first and last frames are fixed. Darker colors indicate later frames. 
    Our method achieves smooth and natural transitions between the conditioned motions.
}
   \label{fig:in_between}
\end{figure*}

\begin{table}[!h]
    \centering
    \scalebox{0.85}{
    \begin{tabular}{cc|cc}
        \toprule
         \multirow{2}{*}{Kernel Size} & \multirow{2}{*}{Norm}  & R Precision&  \multirow{2}{*}{FID $\downarrow$}  \\
           ~& ~ & Top 1$\uparrow$   \\
           
         \midrule
          k=3,1 & BN& \et{0.494}{.005} &\et{5.443}{.093} \\
          k=3,1 & LN & \et{0.488}{.005} &\et{5.339}{.080} \\
          k=3,1 & AdaLN  &\etb{0.501}{.005} &\etb{4.702}{.069} \\
          k=3,3 & AdaLN  & \et{0.484}{.005} &\et{7.120}{.090}\\
          k=5,1 & AdaLN  & \et{0.497}{.005} &\et{5.102}{.064} \\
        \bottomrule
    \end{tabular}
    }
    \caption{\textbf{Ablation studies on LPA.} ``BN'' denotes batch normalization, ``LN'' denotes layer normalization and ``AdaLN'' denotes adaptive layer normalization. ``k=3,1'' means that the first kernel size of the convolution is 3 and the second kernel size is 1.}
    \label{tab:ablations_lpa}
\end{table}

\subsubsection{Design of Localized Pattern Amplification.}
In this section, we conduct ablations on the design of Localized Pattern Amplification (LPA), and the results are shown in~\cref{tab:ablations_lpa}. First, we explore the effect of different normalizations on model performance. Using batch normalization (BN) and layer normalization (LN) does not bring gains for text-to-motion tasks. In contrast, AdaLN, which integrates text information into the extraction process of the local motion pattern, proves to be the optimal normalization method. 
Additionally, we examine how the kernel size of the convolutional layers impacts the model's performance and select the setting of k=3,1.

\begin{figure}[h]
  \centering
   \includegraphics[width=0.82\linewidth]{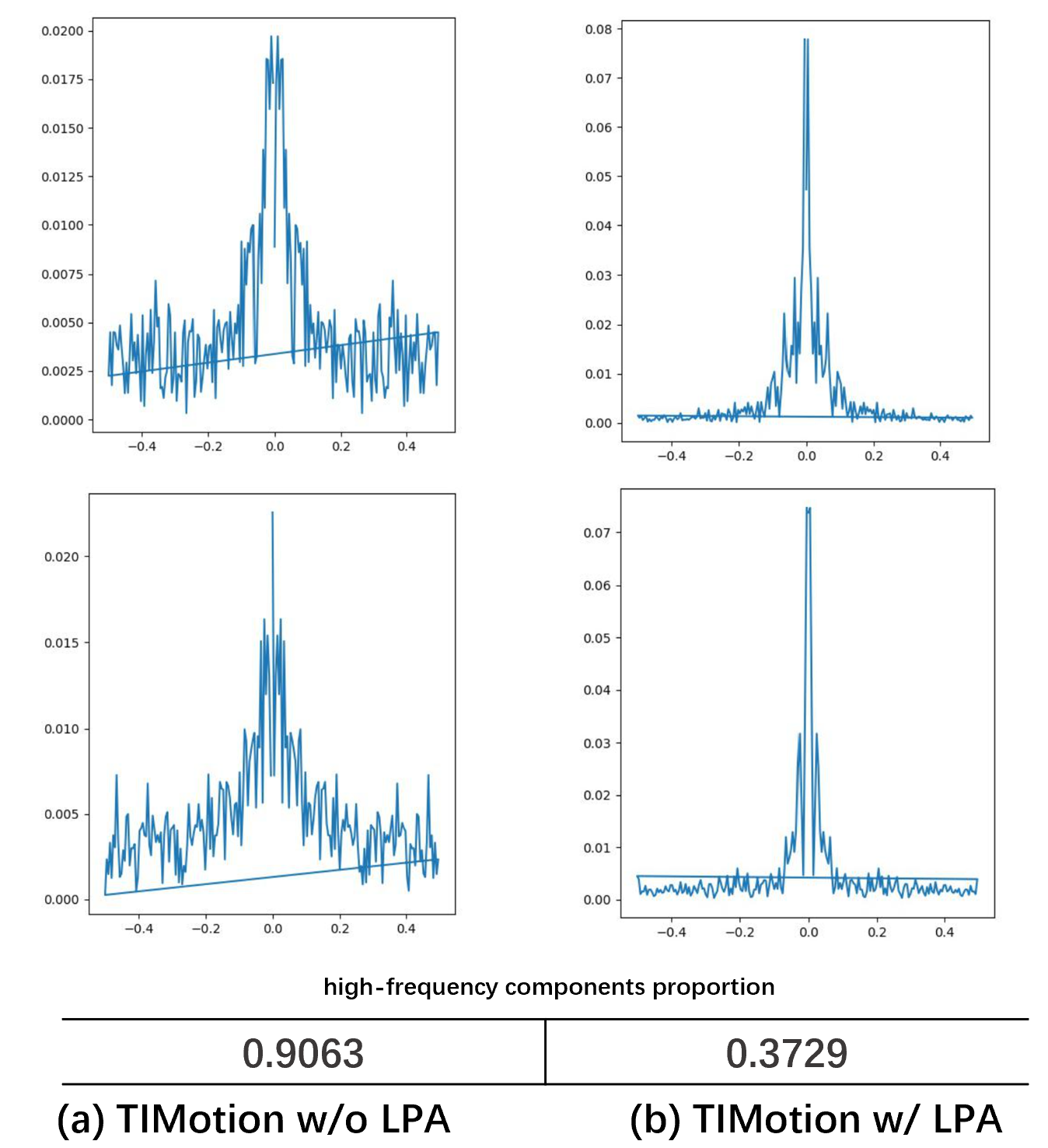}
   \caption{\textbf{Spectrum of motion features.} (a) and (b) show the spectrum of TIMotion w/o and w/ LPA, respectively.
   The horizontal axis denotes the frequency and the vertical axis represents the normalized magnitude. TIMotion w/ LPA contains fewer high-frequency components and therefore generates smoother motion.
}
   \label{fig:lpa}
\end{figure}

To further demonstrate the effectiveness of LPA, we visualize the spectrum of TIMotion w/o and w/ LPA in~\cref{fig:lpa}, respectively. The magnitude has been normalized. As can be seen in~\cref{fig:lpa}, adding LPA reduces the high-frequency component of the feature, making the motion less likely to change drastically and making it smoother.
Moreover, we randomly sampled 200 instances, and w/ LPA, the average proportion of the amplitude of high-frequency components is 0.3729, whereas w/o LPA, the average proportion of the amplitude of high-frequency components is 0.9063.

\subsection{Computational Complexity and Editability.}
\subsubsection{Computational Complexity.}
In~\cref{tab:computation}, we compare our approach TIMotion with the SOTA method InterGen~\cite{liang2024intergen} in terms of computational complexity. TIMotion requires fewer parameters and FLOPs than InterGen but outperforms it on the comprehensive metric FID and R precision. Notably, using a similar transformer architecture to InterGen, TIMotion's average inference time per sample is only 0.632 s while InterGen requires 1.991 s.

\begin{table}[!h]
    \centering
    \scalebox{0.58}{
    \begin{tabular}{c|ccccc}
        \toprule
         \multirow{2}{*}{Methods}  & R Precision&  \multirow{2}{*}{FID $\downarrow$}  & Params &Flops &Inference time\\

           ~ & Top 1$\uparrow$& ~ &(M)  &(G) & (s) \\
       
         \midrule
           InterGen (Transformer) & \et{0.371}{.010} &\et{5.918}{.079}  &182 &80.5 &1.991 \\
          TIMotion+Transformer &\ets{0.491}{.005} &\ets{5.433}{.080}  &\textbf{65} &\textbf{40.4} & \textbf{0.632} \\
          TIMotion+RWKV  &\etb{0.501}{.005} &\etb{4.702}{.069}  &\underline{127} &\underline{58.0} & \underline{1.733}\\
        \bottomrule
    \end{tabular}
    }
    \caption{\textbf{Comparison of computational complexity.} 
    }
    \label{tab:computation}
\end{table}

\subsubsection{Editability.}
\cref{tab:edit} illustrates that TIMotion surpasses InterGen in the task of motion in-betweening editing. We perform experiments on the test set of InterHuman and evaluate by generating 80\% of the sequences based on the first and last 10\% of the sequences.
The quantitative results are shown in~\cref{fig:in_between} and our method achieves smooth and natural transitions between the conditioned motions.

\begin{table}[!h]
    \centering
    \scalebox{0.75}{
    \begin{tabular}{c|cccc}
        \toprule
         \multirow{2}{*}{Methods}  & R Precision&  \multirow{2}{*}{FID $\downarrow$}  & \multirow{2}{*}{MM Dist$\downarrow$} &\multirow{2}{*}{Diversity$\rightarrow$}\\

           ~&   Top 1$\uparrow$  \\
       
         \midrule
           InterGen& \et{0.461}{.006} &\et{4.700}{.066}  &\et{3.780}{.001} &\et{7.682}{.029} \\
          TIMotion& \etb{0.516}{.006} &\etb{3.590}{.049}  &\etb{3.760}{.001} &\etb{7.795}{.031} \\
        \bottomrule
    \end{tabular}
    }
    \caption{Evaluation of motion in-betweening editing task.}
    \label{tab:edit}
\end{table}

\section{Conclusion} \label{sec:conclusion}
In this paper, we abstract the overall human-human motion generation process into a general framework MetaMotion, which consists of two phases: temporal modeling and interaction mixing.
We find that the current methods lead to sub-optimal results and redundancy of model parameters due to inadequate modeling.
Based on this, we introduce TIMotion (Temporal and Interactive Modeling), an efficient and effective approach for human-human motion generation. 
Specifically, we first propose Causal Interactive Injection to model two separate sequences as a causal sequence leveraging the temporal and causal properties.
 Additionally, we proposed Role-Evolving Mixing to adapt to the dynamic roles throughout interactions and designed Localized Pattern Amplification to capture short-term motion patterns for generating smoother and more rational motion.
Extensive experiments on the InterHuman and Inter-X datasets demonstrate that TIMotion significantly outperforms existing methods, achieving state-of-the-art results.

\appendix

\section*{Appendix}

\section{Theoretical Analyses}
We perform gradient magnitude analysis on \emph{separate modeling} ($I$) and our \emph{causal interactive modeling} ($II$). Given that two single-person motion sequences ${X}_a$ and ${X}_b$, the process of separate modeling is:
\begin{equation}
\resizebox{0.7\linewidth}{!}{$
\begin{aligned}
{X}_a^{out} &= {\rm Softmax}(\frac{{X}_a{W}^Q({X}_b{W}^K)^\top}{\sqrt{d}}){X}_b{W}^V, \\
{X}_b^{out} &= {\rm Softmax}(\frac{{X}_b{W}^Q({X}_a{W}^K)^\top}{\sqrt{d}}){X}_a{W}^V,
\end{aligned}
$}
\end{equation}
where ${\rm W^Q}$, ${\rm W^K}$, and ${\rm W^V}$ are trainable weights.
After causal interactive modeling, we can acquire $X$. Then we can obtain the final output as:
\begin{equation}
\resizebox{0.7\linewidth}{!}{$
\begin{aligned}
{X}_{out} &= {\rm Softmax}(\frac{({X}{W}^Q)({X}{W}^K)^\top}{\sqrt{d}}){X}{W}^V.
\end{aligned}
$}
\end{equation}

For ease of analysis, we use the MSE Loss function:
$$
\mathcal{L} = \frac{1}{2}\|{X}_a^{out}-{Y}_a\|_F^2 + \frac{1}{2}\|{X}_b^{out}-{Y}_b\|_F^2
$$
The parameter gradient of separate modeling can be denoted as:
\begin{equation}
\resizebox{0.88\linewidth}{!}{$
\begin{aligned}
\nabla_W \mathcal{L}^{(I)}=\underbrace{\left(\mathbf{X}_a^{\text {out }}-\mathbf{Y}_a\right)}_{\Delta_a} \cdot \frac{\partial \text { CrossAttn }}{\partial W_a}+\underbrace{\left(\mathbf{X}_b^{\text {out }}-\mathbf{Y}_b\right)}_{\Delta_b} \cdot \frac{\partial \text { CrossAttn }}{\partial W_b}.
\end{aligned}
$}
\end{equation}
According to the properties of the matrix 2-norm: $\| A \|_F^2 = \text{Tr}(A^\top A)$, so we can get the F-norm of the parameter gradient as:
\begin{equation}
\resizebox{0.6\linewidth}{!}{$
\begin{aligned}
\|\nabla_W \mathcal{L}^{(I)}\|_F^2 = \|\Delta_a J_a+\Delta_b J_b\|_F^2.
\end{aligned}
$}
\end{equation}
Similarly, we can acquire the parameter gradient of  our causal interactive modeling and its F-norm as:

\begin{equation}
\resizebox{1.0\linewidth}{!}{$
\begin{aligned}
\nabla_W \mathcal{L}^{(I I)} &= \left[\Delta_a \mid \Delta_b\right] \cdot \frac{\partial \text { SelfAttn }}{\partial W} \cdot \mathbf{X}_{o u t}^{\top}, \\
\|\nabla_W \mathcal{L}^{(II)}\|^2 &= \text{Tr}\left( \left( [\Delta_a | \Delta_b] \cdot \frac{\partial \text{SelfAttn}}{\partial W} \cdot X_{out}^\top \right)^\top \left( [\Delta_a | \Delta_b] \cdot \frac{\partial \text{SelfAttn}}{\partial W} \cdot X_{out}^\top \right) \right).
\end{aligned}
$}
\end{equation}
where $\rm Tr$ represents the trace of a matrix. 

As $\left( [\Delta_a | \Delta_b] \cdot J_{self} \cdot X_{out}^\top \right)^\top = X_{out} \cdot J_{self}^\top \cdot [\Delta_a | \Delta_b]^\top$, we get the following equation:
\begin{equation}
\resizebox{0.85\linewidth}{!}{$
\begin{aligned}
\text{Tr}\left( X_{out1} \cdot J_{self}^\top \cdot [\Delta_a | \Delta_b]^\top \cdot [\Delta_a | \Delta_b] \cdot J_{self} \cdot X_{out1}^\top \right)   \\
= \text{Tr}\left( [\Delta_a | \Delta_b]^\top [\Delta_a | \Delta_b] \cdot J_{self} \cdot X_{out1}^\top X_{out1} \cdot J_{self}^\top \right),
\end{aligned}
$}
\end{equation}
where $J$ is the Jacobian matrix.

Assuming the input \( X_{out} \) is normalized and orthogonal ($X_{out}^\top X_{out} = I$), so we can obtain the final results:
\begin{equation}
\resizebox{0.9\linewidth}{!}{$
\begin{aligned}
\left\|\nabla_W \mathcal{L}^{(I I)}\right\|_F^2 &= \operatorname{Tr}\left(\left[\Delta_a \mid \Delta_b\right]^{\top}\left[\Delta_a \mid \Delta_b\right] \cdot J_{s e l f}^{\top} J_{s e l f}\right).
\end{aligned}
$}
\end{equation}

Then we do the following:
\begin{equation}
\resizebox{0.8\linewidth}{!}{$
\begin{aligned}
J_{\text{self}} &= \begin{bmatrix} J_a & J_{ab} \\ J_{ba} & J_b \end{bmatrix}, \\
J_{\text{self}}^\top J_{\text{self}} &= \begin{bmatrix} J_a^\top J_a + J_{ba}^\top J_{ba} & J_a^\top J_{ab} + J_{ba}^\top J_b \\ J_{ab}^\top J_a + J_b^\top J_{ba} & J_{ab}^\top J_{ab} + J_b^\top J_b \end{bmatrix},
\end{aligned}
$}
\end{equation}
Next, we get the following two approximations:
\begin{equation}
\resizebox{0.85\linewidth}{!}{$
\begin{aligned}
J_{\text{self}}^\top J_{\text{self}} = \begin{bmatrix} J_a^\top J_a & J_a^\top J_{ab}  \\ J_{ab}^\top J_a & J_b^\top J_b \end{bmatrix}  + \sigma_{ab}(J_{ab}) + \sigma_{ba}(J_{ba}).
\end{aligned}
$}
\end{equation}

\begin{equation}
\resizebox{1.0\linewidth}{!}{$
\begin{aligned}
\|\nabla_W \mathcal{L}^{(II)}\|_F^2 > \text{Tr}\left( \Delta_a^\top \Delta_a J_a^\top J_a + \Delta_a^\top \Delta_b J_{ab}^\top J_a + \Delta_b^\top \Delta_a J_a^\top J_{ab} + \Delta_b^\top \Delta_b J_b^\top J_b \right)
\end{aligned}
$}
\end{equation}
where \\
\begin{equation}
\resizebox{0.88\linewidth}{!}{$
\begin{aligned}
\text{Tr}(\Delta_a^\top \Delta_a J_a^\top J_a) &= \|\Delta_a J_a\|_F^2, \\
\text{Tr}(\Delta_b^\top \Delta_b J_b^\top J_b) &= \|\Delta_b J_b\|_F^2,  \\
\text{Tr}\left( \Delta_a^\top \Delta_b J_{ab}^\top J_a + \Delta_b^\top \Delta_a J_a^\top J_{ab} \right) &= 2 \cdot \text{Tr}\left( \Delta_a^\top \Delta_b J_{ab}^\top J_a \right).
\end{aligned}
$}
\end{equation}

we apply the Cauchy-Schwarz and get:
\begin{equation}
\resizebox{1.0\linewidth}{!}{$
\begin{aligned}
\|\nabla_W \mathcal{L}^{(II)}\|_F^2 &> \text{Tr}\left( \Delta_a^\top \Delta_a J_a^\top J_a + \Delta_b^\top \Delta_b J_b^\top J_b +  \Delta_a^\top \Delta_b J_{ab}^\top J_a +  \Delta_b^\top \Delta_a J_{a}^\top J_{ab} \right)  \\
&\geq \|\Delta_a J_a+\Delta_b J_b\|_F^2 + 2 \cdot \text{Tr}(\Delta_a^\top \Delta_b J_{ab}^\top J_a) \\
&> \|\Delta_a J_a+\Delta_b J_b\|_F^2 
= \|\nabla_W \mathcal{L}^{(I)}\|_F^2
\end{aligned}.
$}
\end{equation}

Thus, our causal interactive modeling allows for faster model convergence. 

\section{More Experiments on InterX}
To demonstrate the generalizability of our approach TIMotion, we perform corresponding experiments on another large-scale human-human motion generation dataset, InterX~\cite{xu2024interx}. We maintain the same experimental setup as described in the paper. 
The results of comparative methods are directly borrowed from the InterX~\cite{xu2024interx} paper except T2M$^*$~\cite{guo2022t2m} and InterGen$^*$~\cite{liang2024intergen}. The results of T2M$^*$ are taken from the open source repository of InterX~\cite{xu2024interx} and the results of InterGen$^*$ are our own replication based on the unorganized training code provided by the authors of InterX and their open source validation code.
Following established practices~\cite{liang2024intergen}, each experiment is conducted 20 times, and the reported metric values represent the mean with a 95\% statistical confidence interval. The results on InterX are shown in~\cref{tab:main_interX}. In comparison to state-of-the-art (SOTA) approaches, our method, TIMotion, which incorporates various interaction mixing structures (including Transformer, Mamba, and RWKV), consistently outperforms others in terms of FID, R-Precision, Diversity, MM Dist, and MModality.

\begin{table*}[!t]
    \centering
    \scalebox{0.86}{
    \begin{tabular}{l c c c c c c c}
    \toprule
    \multirow{2}{*}{Methods}  & \multicolumn{3}{c}{R Precision$\uparrow$} & \multirow{2}{*}{FID$\downarrow$} & \multirow{2}{*}{MM Dist$\downarrow$} & \multirow{2}{*}{Diversity$\rightarrow$} &\multirow{2}{*}{MModality$\uparrow$}\\

    \cline{2-4}
     ~ & Top 1 & Top 2 & Top 3 \\

    \midrule
    Real & \et{0.429}{.004} & \et{0.626}{.003} & \et{0.736}{.003} & \et{0.002}{.0002} & \et{3.536}{.013} & \et{9.734}{.078} & -  \\ 
    
    \midrule
    
        TEMOS~\cite{petrovich2022temos} & \et{0.092}{.003} & \et{0.171}{.003} & \et{0.238}{.002} & \et{29.258}{.0694} & \et{6.867}{.013} & \et{4.738}{.078} & \et{0.672}{.041} \\   
        T2M~\cite{guo2022t2m} & \et{0.184}{.010} & \et{0.298}{.006} & \et{0.396}{.005} & \et{5.481}{.3280} & \et{9.576}{.006} & \et{5.771}{.151} & \et{2.761}{.042} \\   
        ${\rm T2M^*}$~\cite{guo2022t2m} & \et{0.325}{.004} & \et{0.487}{.005} & \et{0.593}{.005} & \et{3.342}{.0572} & \et{4.506}{.020} & \et{8.535}{.055} & \et{0.982}{.054} \\   
        MDM~\cite{tevet2209mdm} & \et{0.203}{.009} & \et{0.329}{.007} & \et{0.426}{.005} & \et{23.701}{.0569} & \et{9.548}{.014} & \et{5.856}{.077} & \et{3.490}{.061} \\   
        MDM(GRU)~\cite{tevet2209mdm} & \et{0.179}{.006} & \et{0.299}{.005} & \et{0.387}{.007} & \et{32.617}{.1221} & \et{9.557}{.019} & \et{7.003}{.134} & \et{3.430}{.035} \\   
        ComMDM~\cite{shafir2023commdm} & \et{0.090}{.002} & \et{0.165}{.004} & \et{0.236}{.004} & \et{29.266}{.0668} & \et{6.870}{.017} & \et{4.734}{.067} & \et{0.771}{.053} \\   
        InterGen~\cite{liang2024intergen} & \et{0.207}{.004} & \et{0.335}{.005} & \et{0.429}{.005} & \et{5.207}{.2160} & \et{9.580}{.011} & \et{7.788}{.208} & \et{3.686}{.052} \\  
        ${\rm InterGen^*}$~\cite{liang2024intergen} & \et{0.400}{.006} & \et{0.585}{.006} & \et{0.695}{.006} & \et{0.475}{.0305} & \et{3.800}{.020} & \et{9.095}{.055} & \ets{2.657}{.090}  \\
        
    \cline{1-8}
        \textbf{TIMotion+transformer(ours)} & \ets{0.412}{.004} & \ets{0.601}{.004} & \etb{0.714}{.003} & \et{0.385}{.0218} & \etb{3.706}{.015} & \etb{9.191}{.092} & \et{2.437}{.069} \\   
        
        \textbf{TIMotion+mamba(ours)} & \etb{0.414}{.005} & \etb{0.607}{.004} & \ets{0.713}{.003} & \ets{0.348}{.0170} & \etb{3.706}{.014} & \et{9.095}{.058} & \etb{2.779}{.083} \\   
        
        \textbf{TIMotion+RWKV(ours)} & \et{0.411}{.005} & \et{0.597}{.006} & \et{0.707}{.004} & \etb{0.261}{.0140} & \ets{3.737}{.015} & \ets{9.112}{.079} & \et{2.475}{.075}  \\
        
    \bottomrule
    \end{tabular}
    }
    \caption{\textbf{Quantitative evaluation on the InterX~\cite{xu2024interx} test set.} We run the evaluations 20 times. $\pm$ indicates a 95\% confidence interval. \textbf{Bold} indicates the best result, while \underline{underline} refers to the second best. The results of comparative methods are directly borrowed from the InterX~\cite{xu2024interx} paper except T2M$^*$~\cite{guo2022t2m} and InterGen$^*$~\cite{liang2024intergen}. The results of T2M$^*$ are taken from the open source repository of InterX~\cite{xu2024interx} and the results of InterGen$^*$ are our own replication based on the unorganized training code provided by the authors of InterX and their open source validation code.}
    \label{tab:main_interX}

\end{table*}

\section{Algorithm of the motion in-betweening task}
For the motion in-betweening task, we directly use the trained weights from the text-to-motion task.
The overall inference process of motion in-betweening based on TIMotion is shown in~\cref{alg1}.

       

    




   

\begin{algorithm} [h]
    \caption{Inference of TIMotion on the Motion In-betweening Task}
    \label{alg1}
    \begin{flushleft}
    \textbf{Input:} Ground Truth of Motion Sequences for Two Individuals $x^a$ and $x^b$, Length of the Sequence $L$, Ratio of Fixed Sequences $\alpha$, Maximum Timestep of Diffusion $T$, Text Embedding $c$. \\
    \textbf{Output:} Predicted Motion Sequences for Two Individuals $\hat{x}^a_0$ and $\hat{x}^b_0$.
    \end{flushleft}
    \begin{algorithmic}[1]
       
    \STATE $x^a_T \sim \mathcal{N}(0, I)$, \; $x^b_T \sim \mathcal{N}(0, I)$
    
    \FOR{$t = T, \cdots , 1$} 
        \STATE $\hat{x}_0^a, \hat{x}_0^b$ = Diffusion$(x^a_t, x^b_t, t, c)$

        \STATE $\hat{x}_0^a[0:L \cdot \alpha], \hat{x}_0^b[0:L \cdot \alpha]$ = $x^a[0:L \cdot \alpha], x^b[0:L \cdot \alpha]$

        \STATE $\hat{x}_0^a[L-L \cdot \alpha:L], \hat{x}_0^b[L-L \cdot \alpha:L]$ = $x^a[L-L \cdot \alpha:L], x^b[L-L \cdot \alpha:L]$

        \STATE $\epsilon \sim \mathcal{N}(0, I)$

        \STATE $x_{t-1}^a = \sqrt{\overline{\alpha}_{t-1}} \hat{x}_0^a + \sqrt{1- \overline{\alpha}_{t-1}} \epsilon$
        \STATE $x_{t-1}^b = \sqrt{\overline{\alpha}_{t-1}} \hat{x}_0^b + \sqrt{1- \overline{\alpha}_{t-1}} \epsilon$
    \ENDFOR
   
    \end{algorithmic}
\end{algorithm}

       

    




   

\section{More Qualitative Results}
\noindent\textbf{Human-Human motion generation.}
We provide the supplemental demo named \textbf{demo.mp4}.

\noindent\textbf{Motion Editing.}
We provide qualitative results on the motion in-betweening task in~\cref{fig:vis}. Our method achieves smooth and natural transitions between the conditioned and generated motions while complying with the text.

\section{Metrics Computation}
    \textbf{Frechet Inception Distance} (FID): Features are extracted from generated motions and real motions. Subsequently, FID is calculated by comparing the feature distribution of the generated motions with that of the real motions. FID serves as a crucial metric extensively utilized to assess the overall quality of the synthesized motions.
    
    \textbf{R Precision}: For each generated motion, a description pool is created consisting of its ground-truth text description and 31 randomly chosen mismatched descriptions from the test set. Next, the Euclidean distances between the motion and text features of each description in the pool are computed and ranked. We then calculate the average accuracy at the top-1, top-2, and top-3 positions. If the ground truth entry appears among the top-k candidates, it is considered a successful retrieval; otherwise, it is deemed a failure.

    \textbf{MM Dist}: MM distance is calculated as the mean Euclidean distance between the motion feature of each generated motion and the text feature of its corresponding description in the test set.

    \textbf{Diversity}: Diversity measures the variance of the generated motions. From the entire set of generated motions, two subsets of the same size $S_d$ are randomly sampled. Their respective sets of motion feature vectors $\{\mathbf{v}_1,...,\mathbf{v}_{S_d}\}$ and $\{\mathbf{v}_1',...,\mathbf{v}_{S_{d}'}\}$ are extracted. The diversity of this set of motions is defined as
    \begin{equation}
        \mathrm{Diversity} = \frac{1}{S_d}\sum_{i=1}^{S_d}\parallel \mathbf{v}_i-\mathbf{v}_i' \parallel_2.
    \end{equation}
    $S_d=300$ is used in experiments.
    
    \textbf{MModality}: MModality measures how much the generated motions diversify within the same text. Given a set of motions corresponding to a specific text,  two subsets of the same size $S_l$ are randomly sampled. Their respective sets of motion feature vectors $\{\mathbf{v}_{c,1},...$ $,\mathbf{v}_{c, S_l}\}$ and $\{\mathbf{v}_{c,1}',...,\mathbf{v}_{c, S_l}'\}$ are extracted. The MModality of this motion set is formalized as
    \begin{equation}
        \mathrm{Multimodality} = \frac{1}{C \times S_l} \sum_{c=1}^C \sum_{i=1}^{S_l} \left\| \mathbf{v}_{c,i}-\mathbf{v}'_{c,i} \right\|_2.
    \end{equation}
    ${S_l}=100$ is used in experiments.

\section{Limitation}
Limited by the variety of two-person datasets, we demonstrated the effectiveness of TIMotion mainly on the text-to-motion task. In the future, we will validate our method on more tasks based on newly released datasets.
Moreover, our proposed TIMotion effectively models the motion relationship between the two individuals, but the modeling of motion relationships between three or more people has not been explored yet. Related researchers in the community are encouraged to explore more on motion modeling among people and TIMoiton may provide some new insight for the community. 

{
    \small
    \bibliographystyle{ieeenat_fullname}
    \bibliography{main}
}


{
    \small
    \bibliographystyle{ieeenat_fullname}
    \bibliography{main}

@String(TOG= {ACM Trans. Graph.})

@String(ICLR = {Int. Conf. Learn. Represent.})

@String(TOG   = {ACM TOG})

@String(ICLR  = {ICLR})

@inproceedings{petrovich2022temos,
  title={TEMOS: Generating diverse human motions from textual descriptions},
  author={Petrovich, Mathis and Black, Michael J and Varol, G{\"u}l},
  booktitle={European Conference on Computer Vision},
  pages={480--497},
  year={2022},
  organization={Springer}
}

@inproceedings{guo2022t2m,
  title={Generating diverse and natural 3d human motions from text},
  author={Guo, Chuan and Zou, Shihao and Zuo, Xinxin and Wang, Sen and Ji, Wei and Li, Xingyu and Cheng, Li},
  booktitle={Proceedings of the IEEE/CVF Conference on Computer Vision and Pattern Recognition},
  pages={5152--5161},
  year={2022}
}

@article{zhang2024motiondiffuse,
  title={Motiondiffuse: Text-driven human motion generation with diffusion model},
  author={Zhang, Mingyuan and Cai, Zhongang and Pan, Liang and Hong, Fangzhou and Guo, Xinying and Yang, Lei and Liu, Ziwei},
  journal={IEEE Transactions on Pattern Analysis and Machine Intelligence},
  year={2024},
  publisher={IEEE}
}

@article{tevet2209mdm,
  title={Human motion diffusion model},
  author={Tevet, G and Raab, S and Gordon, B and Shafir, Y and Cohen-Or, D and Bermano, AH},
  journal={arXiv preprint arXiv:2209.14916},
  year={2022}
}

@article{shafir2023commdm,
  title={Human motion diffusion as a generative prior},
  author={Shafir, Yonatan and Tevet, Guy and Kapon, Roy and Bermano, Amit H},
  journal={arXiv preprint arXiv:2303.01418},
  year={2023}
}

@inproceedings{tanaka2023rig,
  title={Role-aware interaction generation from textual description},
  author={Tanaka, Mikihiro and Fujiwara, Kent},
  booktitle={Proceedings of the IEEE/CVF international conference on computer vision},
  pages={15999--16009},
  year={2023}
}

@article{liang2024intergen,
  title={Intergen: Diffusion-based multi-human motion generation under complex interactions},
  author={Liang, Han and Zhang, Wenqian and Li, Wenxuan and Yu, Jingyi and Xu, Lan},
  journal={International Journal of Computer Vision},
  pages={1--21},
  year={2024},
  publisher={Springer}
}

@inproceedings{xu2024interx,
  title={Inter-x: Towards versatile human-human interaction analysis},
  author={Xu, Liang and Lv, Xintao and Yan, Yichao and Jin, Xin and Wu, Shuwen and Xu, Congsheng and Liu, Yifan and Zhou, Yizhou and Rao, Fengyun and Sheng, Xingdong and others},
  booktitle={Proceedings of the IEEE/CVF Conference on Computer Vision and Pattern Recognition},
  pages={22260--22271},
  year={2024}
}

@article{song2020diffusion,
  title={Denoising diffusion implicit models},
  author={Song, Jiaming and Meng, Chenlin and Ermon, Stefano},
  journal={arXiv preprint arXiv:2010.02502},
  year={2020}
}

@inproceedings{nichol2021cosnoise,
  title={Improved denoising diffusion probabilistic models},
  author={Nichol, Alexander Quinn and Dhariwal, Prafulla},
  booktitle={International conference on machine learning},
  pages={8162--8171},
  year={2021},
  organization={PMLR}
}

@article{ho2022classifier,
  title={Classifier-free diffusion guidance},
  author={Ho, Jonathan and Salimans, Tim},
  journal={arXiv preprint arXiv:2207.12598},
  year={2022}
}

@article{loshchilov2017adam,
  title={Decoupled weight decay regularization},
  author={Loshchilov, Ilya and Hutter, Frank},
  journal={arXiv preprint arXiv:1711.05101},
  year={2017}
}

@article{ho2020denoising,
  title={Denoising diffusion probabilistic models},
  author={Ho, Jonathan and Jain, Ajay and Abbeel, Pieter},
  journal={Advances in neural information processing systems},
  volume={33},
  pages={6840--6851},
  year={2020}
}

@inproceedings{radford202clip,
  title={Learning transferable visual models from natural language supervision},
  author={Radford, Alec and Kim, Jong Wook and Hallacy, Chris and Ramesh, Aditya and Goh, Gabriel and Agarwal, Sandhini and Sastry, Girish and Askell, Amanda and Mishkin, Pamela and Clark, Jack and others},
  booktitle={International conference on machine learning},
  pages={8748--8763},
  year={2021},
  organization={PMLR}
}

@inproceedings{tevet2022motionclip,
  title={Motionclip: Exposing human motion generation to clip space},
  author={Tevet, Guy and Gordon, Brian and Hertz, Amir and Bermano, Amit H and Cohen-Or, Daniel},
  booktitle={European Conference on Computer Vision},
  pages={358--374},
  year={2022},
  organization={Springer}
}

@inproceedings{zhong2023attt2m,
  title={Attt2m: Text-driven human motion generation with multi-perspective attention mechanism},
  author={Zhong, Chongyang and Hu, Lei and Zhang, Zihao and Xia, Shihong},
  booktitle={Proceedings of the IEEE/CVF International Conference on Computer Vision},
  pages={509--519},
  year={2023}
}

@inproceedings{gong2023tm2d,
  title={Tm2d: Bimodality driven 3d dance generation via music-text integration},
  author={Gong, Kehong and Lian, Dongze and Chang, Heng and Guo, Chuan and Jiang, Zihang and Zuo, Xinxin and Mi, Michael Bi and Wang, Xinchao},
  booktitle={Proceedings of the IEEE/CVF International Conference on Computer Vision},
  pages={9942--9952},
  year={2023}
}

@article{van2017vqvae,
  title={Neural discrete representation learning},
  author={Van Den Oord, Aaron and Vinyals, Oriol and others},
  journal={Advances in neural information processing systems},
  volume={30},
  year={2017}
}

@inproceedings{zhang2023t2mgpt,
  title={Generating human motion from textual descriptions with discrete representations},
  author={Zhang, Jianrong and Zhang, Yangsong and Cun, Xiaodong and Zhang, Yong and Zhao, Hongwei and Lu, Hongtao and Shen, Xi and Shan, Ying},
  booktitle={Proceedings of the IEEE/CVF conference on computer vision and pattern recognition},
  pages={14730--14740},
  year={2023}
}

@inproceedings{chen2023mld,
  title={Executing your commands via motion diffusion in latent space},
  author={Chen, Xin and Jiang, Biao and Liu, Wen and Huang, Zilong and Fu, Bin and Chen, Tao and Yu, Gang},
  booktitle={Proceedings of the IEEE/CVF Conference on Computer Vision and Pattern Recognition},
  pages={18000--18010},
  year={2023}
}

@inproceedings{zhang2023remodiffuse,
  title={Remodiffuse: Retrieval-augmented motion diffusion model},
  author={Zhang, Mingyuan and Guo, Xinying and Pan, Liang and Cai, Zhongang and Hong, Fangzhou and Li, Huirong and Yang, Lei and Liu, Ziwei},
  booktitle={Proceedings of the IEEE/CVF International Conference on Computer Vision},
  pages={364--373},
  year={2023}
}

@article{wang2023intercontrol,
  title={Intercontrol: Generate human motion interactions by controlling every joint},
  author={Wang, Zhenzhi and Wang, Jingbo and Lin, Dahua and Dai, Bo},
  journal={arXiv preprint arXiv:2311.15864},
  year={2023}
}

@article{peng2023rwkv,
  title={Rwkv: Reinventing rnns for the transformer era},
  author={Peng, Bo and Alcaide, Eric and Anthony, Quentin and Albalak, Alon and Arcadinho, Samuel and Biderman, Stella and Cao, Huanqi and Cheng, Xin and Chung, Michael and Grella, Matteo and others},
  journal={arXiv preprint arXiv:2305.13048},
  year={2023}
}

@article{gu2023mamba,
  title={Mamba: Linear-time sequence modeling with selective state spaces},
  author={Gu, Albert and Dao, Tri},
  journal={arXiv preprint arXiv:2312.00752},
  year={2023}
}

@inproceedings{vaswani2017transformer,
  title={Attention is all you need},
  author={Ashish Vaswani and Noam Shazeer and
                  Niki Parmar and
                  Jakob Uszkoreit and
                  Llion Jones and
                  Aidan N. Gomez and
                  Lukasz Kaiser and
                  Illia Polosukhin},
  booktitle={Advances in Neural Information Processing Systems},
  year={2017}
}

@book{parent2012computer,
  title={Computer animation: algorithms and techniques},
  author={Parent, Rick},
  year={2012},
  publisher={Newnes}
}

@book{magnenat1985computer,
  title={Computer animation},
  author={Magnenat-Thalmann, Nadia and Thalmann, Daniel and Magnenat-Thalmann, Nadia and Thalmann, Daniel},
  year={1985},
  publisher={Springer}
}

@article{urbain2010introduction,
  title={Introduction to game development},
  author={Urbain, Jay},
  journal={Cell},
  volume={414},
  pages={745--5102},
  year={2010}
}

@book{bethke2003game,
  title={Game development and production},
  author={Bethke, Erik},
  year={2003},
  publisher={Wordware Publishing, Inc.}
}

@article{saridis1983intelligent,
  title={Intelligent robotic control},
  author={Saridis, George},
  journal={IEEE Transactions on Automatic Control},
  volume={28},
  number={5},
  pages={547--557},
  year={1983},
  publisher={IEEE}
}

@inproceedings{wang2023towards,
  title={Towards domain generalization for multi-view 3d object detection in bird-eye-view},
  author={Wang, Shuo and Zhao, Xinhai and Xu, Hai-Ming and Chen, Zehui and Yu, Dameng and Chang, Jiahao and Yang, Zhen and Zhao, Feng},
  booktitle={Proceedings of the IEEE/CVF Conference on Computer Vision and Pattern Recognition},
  pages={13333--13342},
  year={2023}
}

@inproceedings{guo2020action2motion,
  title={Action2motion: Conditioned generation of 3d human motions},
  author={Guo, Chuan and Zuo, Xinxin and Wang, Sen and Zou, Shihao and Sun, Qingyao and Deng, Annan and Gong, Minglun and Cheng, Li},
  booktitle={Proceedings of the 28th ACM International Conference on Multimedia},
  pages={2021--2029},
  year={2020}
}

@inproceedings{petrovich2021action,
  title={Action-conditioned 3d human motion synthesis with transformer vae},
  author={Petrovich, Mathis and Black, Michael J and Varol, G{\"u}l},
  booktitle={Proceedings of the IEEE/CVF International Conference on Computer Vision},
  pages={10985--10995},
  year={2021}
}

@inproceedings{pinyoanuntapong2024mmm,
  title={Mmm: Generative masked motion model},
  author={Pinyoanuntapong, Ekkasit and Wang, Pu and Lee, Minwoo and Chen, Chen},
  booktitle={Proceedings of the IEEE/CVF Conference on Computer Vision and Pattern Recognition},
  pages={1546--1555},
  year={2024}
}

@inproceedings{guo2024momask,
  title={Momask: Generative masked modeling of 3d human motions},
  author={Guo, Chuan and Mu, Yuxuan and Javed, Muhammad Gohar and Wang, Sen and Cheng, Li},
  booktitle={Proceedings of the IEEE/CVF Conference on Computer Vision and Pattern Recognition},
  pages={1900--1910},
  year={2024}
}

@inproceedings{habibie2022speech,
  title={A motion matching-based framework for controllable gesture synthesis from speech},
  author={Habibie, Ikhsanul and Elgharib, Mohamed and Sarkar, Kripasindhu and Abdullah, Ahsan and Nyatsanga, Simbarashe and Neff, Michael and Theobalt, Christian},
  booktitle={ACM SIGGRAPH 2022 conference proceedings},
  pages={1--9},
  year={2022}
}

@article{ao2022speech,
  title={Rhythmic gesticulator: Rhythm-aware co-speech gesture synthesis with hierarchical neural embeddings},
  author={Ao, Tenglong and Gao, Qingzhe and Lou, Yuke and Chen, Baoquan and Liu, Libin},
  journal={ACM Transactions on Graphics (TOG)},
  volume={41},
  number={6},
  pages={1--19},
  year={2022},
  publisher={ACM New York, NY, USA}
}

@inproceedings{cai2024digital,
  title={Digital life project: Autonomous 3d characters with social intelligence},
  author={Cai, Zhongang and Jiang, Jianping and Qing, Zhongfei and Guo, Xinying and Zhang, Mingyuan and Lin, Zhengyu and Mei, Haiyi and Wei, Chen and Wang, Ruisi and Yin, Wanqi and others},
  booktitle={Proceedings of the IEEE/CVF Conference on Computer Vision and Pattern Recognition},
  pages={582--592},
  year={2024}
}

@article{he2024mambaad,
  title={Mambaad: Exploring state space models for multi-class unsupervised anomaly detection},
  author={He, Haoyang and Bai, Yuhu and Zhang, Jiangning and He, Qingdong and Chen, Hongxu and Gan, Zhenye and Wang, Chengjie and Li, Xiangtai and Tian, Guanzhong and Xie, Lei},
  journal={arXiv preprint arXiv:2404.06564},
  year={2024}
}

@inproceedings{GuyTevet_MDM,
  title={Human Motion Diffusion Model},
  author={Guy Tevet and Sigal Raab and Brian Gordon and Yonatan Shafir and Daniel Cohen{-}Or and Amit Haim Bermano},
  booktitle={ICLR},
  year={2023}
}

@inproceedings{zhao2024wavelet,
  title={Wavelet-based fourier information interaction with frequency diffusion adjustment for underwater image restoration},
  author={Zhao, Chen and Cai, Weiling and Dong, Chenyu and Hu, Chengwei},
  booktitle={Proceedings of the IEEE/CVF Conference on Computer Vision and Pattern Recognition},
  pages={8281--8291},
  year={2024}
}

@inproceedings{fan2024freemotion,
  title={Freemotion: A unified framework for number-free text-to-motion synthesis},
  author={Fan, Ke and Tang, Junshu and Cao, Weijian and Yi, Ran and Li, Moran and Gong, Jingyu and Zhang, Jiangning and Wang, Yabiao and Wang, Chengjie and Ma, Lizhuang},
  booktitle={European Conference on Computer Vision},
  pages={93--109},
  year={2024},
  organization={Springer}
}

@inproceedings{wang2024sqdmap,
  title={Stream Query Denoising for Vectorized HD-Map Construction},
  author={Wang, Shuo and Jia, Fan and Mao, Weixin and Liu, Yingfei and Zhao, Yucheng and Chen, Zehui and Wang, Tiancai and Zhang, Chi and Zhang, Xiangyu and Zhao, Feng},
  booktitle={European Conference on Computer Vision},
  pages={203--220},
  year={2024},
  organization={Springer}
}

@article{achiam2023gpt,
  title={Gpt-4 technical report},
  author={Achiam, Josh and Adler, Steven and Agarwal, Sandhini and Ahmad, Lama and Akkaya, Ilge and Aleman, Florencia Leoni and Almeida, Diogo and Altenschmidt, Janko and Altman, Sam and Anadkat, Shyamal and others},
  journal={arXiv preprint arXiv:2303.08774},
  year={2023}
}
}


\end{document}